\pdfoutput=1

\documentclass[11pt]{article}

\usepackage[]{acl}

\usepackage{times}
\usepackage{latexsym}

\usepackage[T1]{fontenc}

\usepackage[utf8]{inputenc}

\usepackage{microtype}

\usepackage{inconsolata}

\usepackage[linesnumbered,ruled,vlined]{algorithm2e}
\usepackage{graphicx}
\usepackage{amsmath,amsfonts,bm}
\usepackage{multirow}

\usepackage{enumitem}
\usepackage{graphicx}
\graphicspath{figures/}
\usepackage{subfig}
\usepackage{multirow}
\usepackage{extarrows}
\usepackage{acronym}
\usepackage{xspace}
\usepackage[capitalize]{cleveref}
\usepackage{caption}

\usepackage{wrapfig}
\usepackage{booktabs}
\usepackage{titletoc}
\usepackage{appendix}
\usepackage{pifont}

\makeatletter
\DeclareRobustCommand\onedot{\futurelet\@let@token\@onedot}
\def\@onedot{\ifx\@let@token.\else.\null\fi\xspace}

\def\ie{\emph{i.e}\onedot}

\makeatother

\usepackage{algpseudocode}
\usepackage{bbm}
\usepackage{float}
\usepackage{xcolor,colortbl}
\usepackage{makecell}

\usepackage{diagbox}
\usepackage{verbatim}
\usepackage{framed}
\usepackage[most]{tcolorbox}


\title{Combining Supervised Learning and Reinforcement Learning for Multi-Label Classification Tasks with Partial Labels}

\author{Zixia Jia$^{1,2}$, Junpeng Li$^{1,2}$, Shichuan Zhang$^3$, Anji Liu$^4$, Zilong Zheng$^{1,2}$\Thanks{~Correspondence to Zilong Zheng.} \\
$^1$ Beijing Institute for General Artificial Intelligence (BIGAI), Beijing, China \\
$^2$ National Key Labor
$^3$ Zhejiang University, Hangzhou, Zhejiang, China
$^4$ UCLA
\\
\texttt{ \{jiazixia,zlzheng\}@bigai.ai,  jpl\_xd@163.com} \\ 
\texttt{  zhangshichuan@westlake.edu.cn, liuanji@cs.ucla.edu}}

\begin{document}
\maketitle
\begin{abstract}

Traditional supervised learning heavily relies on human-annotated datasets, especially in data-hungry neural approaches. However, various tasks, especially multi-label tasks like document-level relation extraction, pose challenges in fully manual annotation due to the specific domain knowledge and large class sets. Therefore, we address the multi-label positive-unlabelled learning (MLPUL) problem, where only a subset of positive classes is annotated. We propose Mixture Learner for Partially Annotated Classification (MLPAC), an RL-based framework combining the exploration ability of reinforcement learning and the exploitation ability of supervised learning. Experimental results across various tasks, including document-level relation extraction, multi-label image classification, and binary PU learning, demonstrate the generalization and effectiveness of our framework.
\end{abstract}

\section{Introduction}

\begin{figure*}[t!]
\centering
\small
  \includegraphics[width=\linewidth]{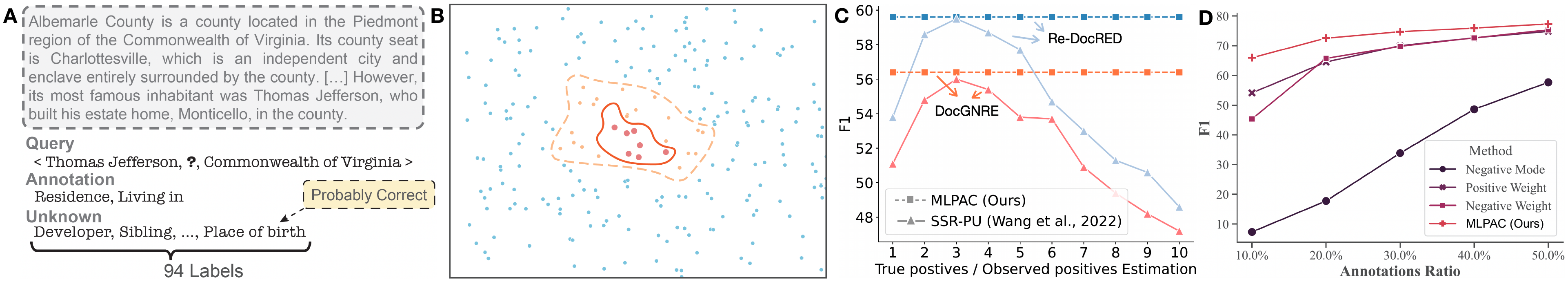}
  \caption{\textbf{A.} A partially annotated data sample in DocRE task. \textbf{B.} Severe imbalanced distribution of annotated positive (red scatters) and negative labels (blue and orange scatters) corresponding to the DocRED \cite{yao-etal-2019-docred} dataset. Orange scatters are actually-positive labels reannotated by Re-DocRED dataset \cite{wu2022revisiting}  
  \textbf{C.} Results of training on incomplete DocRED and testing on reannotated Re-DocRED and DocGNRE \cite{li2023semi}. SSR-PU is sensitive to prior estimation, while ours is prior agnostic.
  \textbf{D.} Performance comparison in DocRE task.} 
  \label{fig:intro}
 
\end{figure*}

Multi-Label Classification (MLC) task treats a problem that allows instances to take multiple labels, and traditional Supervised Learning (SL) methods on MLC heavily rely on human-annotated data sets, especially neural approaches that are data-hungry and susceptible to over-fitting when lacking training data. However, in many MLC tasks that generally have dozens or hundreds of sizes of class sets, incompleteness in the acquired annotations frequently arises owing to the limited availability of expert annotators or the subjective nature inherent in human annotation processes.
\cite{kanehira2016multi, cole2021multi, tan2022revisiting, ben2022multi}. Therefore, we focus on the fundamentally important problem, typically termed Multi-Label Positive-Unlabelled Learning (MLPUL) \cite{kanehira2016multi, teisseyre2021classifier}, which involves learning from a multi-label dataset in which only \textit{a subset of positive classes} is definitely annotated, while all the remaining classes are unknown (which could be positives or negatives). For instance, as shown in \cref{fig:intro}A\&B, human annotators find it hard to completely annotate all the relations due to the confusion of understanding relation definitions and long-context semantics in document-level relation extraction (DocRE) task \cite{huang2022does, tan2022revisiting}.   

Positive and unlabelled (PU) classification has received extensive attention in binary settings, with several recent MLPUL approaches adapting traditional binary PU loss functions to address multi-label classification tasks \cite{kanehira2016multi, wang2022unified, wang2023positive}. These methods typically operate under the assumption that the prior distribution of positive labels can be inferred from fully labeled training samples or closely unbiased estimations. In a specific instance, \cite{wang2022unified} supposed that the actual positive classes are three times the number of observed labels 
in the DocRE task, and their model's performance is heavily influenced by the prior (\cref{fig:intro}C). 
However, estimating the prior distribution of labels in real-world scenarios poses significant challenges, as it is rarely feasible to ensure a comprehensive data set encompassing all label types
\cite{chen2020variational,hu2021predictive, yuan2023positive}.
Additionally, \citet{li2023semi} noted that many long-tail label types tend to be omitted from training annotations. 
Consequently, we focus on addressing MLPUL without prior knowledge of class distribution. 
Moreover, MLC generally faces the challenge of imbalanced positive and negative labels, which is severely exacerbated by missing positive class annotations under MLPUL, as shown in \cref{fig:intro}B. Previous works typically adopted the re-balance factor to re-weight the loss functions, containing positive up-weight and negative under-weight \citep{li2020empirical}. We simply attempt these approaches and find they partly improve the model performance but still perform unsatisfactorily when only a very small set ($10\%$) of positive class annotations is available (\cref{fig:intro}D). 

Previous works \citep{silver2016mastering, feng2018reinforcement, nooralahzadeh2019reinforcement} have demonstrated the powerful exploration ability of Reinforcement Learning (RL). Furthermore, RL has shown great success on distant or partial annotations recently \cite{
feng2018reinforcement, 
luo2021pulns, chen2023reinforcement}. 
Inspired by these successful RL attempts, we believe that the exploratory nature of RL has the potential ability to discover additional positive classes while mitigating the overfitting issues typically encountered in supervised learning, especially when the observed label distribution is severely biased, which holds promise in addressing MLPUL.
Besides, recent works have shown that supervised learning can be remarkably effective for the RL process \cite{emmons2021rvs, park2021surf, badrinath2023waypoint}. 

Based on this intuition, we introduce a novel framework termed Mixture Learner for Partially Annotated Classification (MLPAC), which combines the exploratory capacity of RL in tandem with the exploitation capabilities of supervised learning.
Specifically, we design a policy network (as a multi-label classifier) and a critic network, along with two types of reward functions: global rewards calculated by a recall function, which evaluates the all-classes prediction performance for each instance, and local rewards provided by the critic network, which assesses the prediction quality of each individual class for a given instance.
The local rewards are expected to narrow the exploration space of traditional RL and offer a preliminary yet instructive signal to guide the learning process,
while the global rewards encourage the policy network to explore a broader spectrum of positive classes, consequently mitigating distribution bias stemming from imbalanced labels and incomplete annotations.

In addition, inspired by the traditional actor-critic RL algorithm \citep{bahdanau2016actor}, we iteratively train the policy network and the critic network, which achieves dynamic reward estimation in our setting.
The absence of fully annotated samples in both training and validation sets precludes 
the direct attainment of perfectly accurate rewards. Hence, we introduce label enhancement through collaborative policy and critic network efforts during iterative training, boosting label confidence and enhancing the critic network's reward estimation accuracy.

Moreover, our RL framework is concise and flexible, guaranteeing its generalization and adaptation to many tasks. 
Beyond the experiments on document-level relation extraction task (\S\ref{sec:doc_exp}) in Natural Language Process (NLP) field,
we also conduct sufficient experiments in multi-label image classification task~(\S\ref{sec:image_exp}) in Computer Vision (CV) field and general PU learning setting in binary case (\S\ref{sec:toy_exp}) to verify the generalization and effectiveness of our framework. All experimental results demonstrate the advantage and significant improvement of our framework.

\begin{figure*}[t!]
\centering
\small
\includegraphics[width=.9\linewidth]{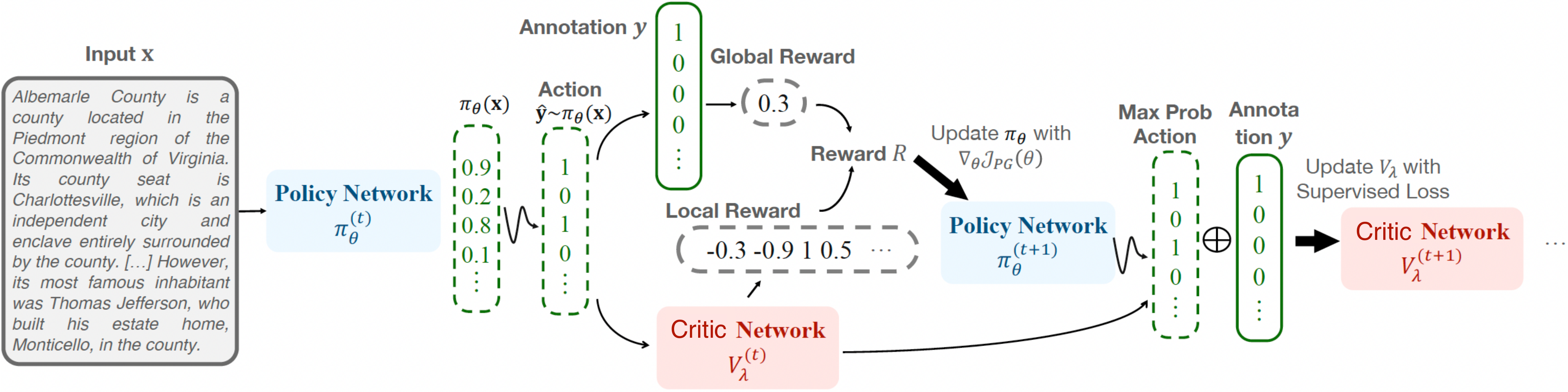}
    \caption{Illustration of our RL framework.  $\bigoplus$ represents union operation. We iteratively update the policy network and critic network. The augmented training data are curated for the critic network.}
    \label{fig:teaser}

\end{figure*}

\section{Related Work}

\paragraph{Multi-label Positive-Unlabelled Learning Methods} 
Label correlation modeling, rank-based
weighted loss function and enhanced incomplete labels are typical technologies that previous MLPUL methods adopted. Label correlation is usually learned or calculated from the label matrix of the training data under partial positive \textit{and} negative samples under \textit{multi-label partially observed labeling} \cite{rastogi2021multi, kumar2022low, jiang2023global, yu2024enhancing}. 
In the MLPUL setting, \citet{teisseyre2021classifier} leveraged classifier chains to model high-order label correlation with the assumption that prior probabilities could be estimated by predictions from the previous models in the chain. Because label correlations may bring correlation bias under the severe scarcity of positive labels, we did not explicitly model the correlations in our model. 

\citet{kanehira2016multi} extended binary PU classification to the multi-label setting, dealing with multi-label PU ranking problem. They modeled MLPUL as cost-sensitive learning by designing a weighted rank loss for multi-label image classification tasks. Following this, \citet{wang2022unified} proposed shift and squared ranking loss PU learning for the document-level relation extraction task by modifying the prior terms in the rank-based loss function. Both works supposed the knowledge of class prior distribution, but the prior is difficult to estimate in reality.
Recently, encouraged by the binary PU methodology without class prior \cite{chen2020variational,hu2021predictive},
\citet{yuan2023positive} adopted the variational binary PU loss and additionally dealt with the catastrophic imbalanced positive and negative label distribution that MLPUL faced by introducing an adaptive re-balance factor and adaptive temperature coefficient in the loss function. Our approach also pays attention to solving the imbalanced problem in MLPUL and designs a framework without the knowledge of class priors. Different from their methods specifically designed for image classification, our framework is effective for a wide range of MLPUL tasks and is further adapted for imbalanced binary PU learning cases.

Besides the above methodologies, \citet{chen2020variational, yuan2023positive} proposed regularization term based on Mixup \cite{zhang2017mixup} to enhance incomplete labels, alleviating the overfitting problem and increasing model robustness. The concurrent work to us, \citet{wang2023positive} designs positive-augmentation and positive-mixup strategies to improve rank-based learning methods \cite{wang2022unified} for PU document-level relation extraction. These strategies are not uniquely applicable to their framework, and our method can also integrate them, which we leave to future work.

\paragraph{Reinforcement Learning under Weak Supervision}

There are many previous works leveraging Reinforcement Learning (RL) to solve tasks only with weak supervision \citep{feng2018reinforcement, zeng2018large, luo2021pulns, chen2023reinforcement}. In the NLP field, to precisely leverage distant data, \citet{qin2018robust, feng2018reinforcement} train an agent as a noisy-sentence filter, taking performance variation on development or probabilities of selected samples as a reward and adopting policy gradient to update. \citet{nooralahzadeh2019reinforcement} expand their methods to NER task. Recent work of \citet{chen2023reinforcement} also conducts RL to remove the noisy sentence so as to improve the fault diagnosis system. Notably, 
RL learning on distantly supervised learning aims to filter false positives, whereas our goal is to identify false negatives. A closer related work to us is \citet{luo2021pulns}, in which an RL method is designed to solve the PU learning problem. But unlike us, their agent is a \textit{negative sample selector}, aiming to find negatives with high confidence and then couple them with partial positives to train a classifier. 
More related works under different weak supervision settings can be found in Appendix \ref{appd:related}.

\section{RL-based Framework}

We propose a novel RL framework to solve the MLPUL task. We formulate the multi-label prediction as the action execution in the Markov Decision Process (MDP) \citep{puterman1990markov}. We design both \textit{local} and \textit{global} rewards for actions to guide the action decision process. The policy-based RL method is adopted to train our policy network. The overall RL framework is illustrated in \Cref{fig:teaser}. 
We introduce the details in the following subsections. 

\subsection{Problem Setting}
\label{sec:problem_set}
We mathematically formulate the MLPUL task. Given a multi-label dataset $\mathcal{X} = \{\mathbf{x}_i\}$, each $\mathbf{x}_i$ is labeled with a partially annotated multi-hot vector $\mathbf{y}_i = [y^1_i, \cdots, y^c_i, \cdots y^{|\mathcal{C}|}_i]$, where $y^c_i \in\{0,1\}$ denotes whether class $c \in \mathcal{C}$ is \texttt{TRUE} for instance $\mathbf{x}_i$ and $|\mathcal{C}|$ is the cardinality of set $\mathcal{C}$. In terms of partial annotation, we assume that $\{c; y_i^c = 1\}$ is a subset of gold positive classes regarding to $\mathbf{x}_i$, \ie, $\{c ; y_i^c = 0\}$ is the set of \texttt{UNKNOWN} classes (which could be actually positive or negative). Typically, in the multi-label setting, the size of the label set is dozens or hundreds; thus $|\{c; y_i^c = 1\}| << |\{c; y_i^c = 0\}| \leq |\mathcal{C}|$. 
In some complicated tasks, such as Relation Extraction, we further define a special label \texttt{<None>} to $\mathbf{x}_i$ if $\mathbf{y}_i = [y_i^c = 0]_{c=1}^{|\mathcal{C}|}$. It should be mentioned that we do not have any fully annotated data, both the training and validation sets being partially annotated.


A straightforward approach (referred to as the ``negative mode'') for tackling this task involves treating all unlabeled classes as \texttt{FALSE} (set all $y^c_i = 0$ to $y^c_i=-1$) and subsequently reducing the multi-label problem to a number of independent binary classiﬁcation tasks by employing conventional supervised learning (SL). However, due to incomplete positive labels and severely imbalanced label distribution, the negative mode is susceptible to overfit biased annotated labels, resulting in high precision but low recall on the test set. Based on the negative mode, we introduce an RL framework (MLPAC) that combines RL and SL to mitigate distribution bias and encourage the multi-label classifier to predict more potential positive classes.

\subsection{Modeling}

Typically, basic RL is modeled as an MDP $(S, A, \pi, \mathcal{T}, R)$ which contains a set of environment and agent states $S$, a set of actions $A$ of the agent, the transition probabilities $\mathcal{T}$ from a state to another state under action $a$, and the reward $R$. The goal of an RL agent is to learn a policy $\pi$ that maximizes the expected cumulative reward. In our problem setting, we formalize the multi-label positive-unlabelled learning as a \textbf{one-step} MDP problem: we do not consider state transitions because our action execution does not change the agent and environment. Our setting highly resembles the setting of contextual bandits \cite{chu2011contextual}, where actions only affect the reward but not the state. Our RL framework's policy $\pi_\theta$ is a multi-label classifier constructed by a neural network with parameter $\theta$. We define the constituents 
in detail.

\paragraph{States}  
A state $s$ includes the potential information of an instance to be labeled. In our setting, this information consists of instance features, which are essentially continuous real-valued vectors derived from a neural network. 

\paragraph{Actions} Due to the multi-label setting, our agent is required to determine the label of each class $c$ for one instance. There are two actions for our agent: setting the current class as \texttt{TRUE} ($\hat{y}^c_{i}=1$) or \texttt{FALSE} ($\hat{y}^c_{i}=-1$). It is necessary to execute $|\mathcal{C}|$ (size of the class set) actions to label an instance completely. 

\paragraph{Policy} Our policy network outputs the probability $\pi_\theta(\hat{y}^c_{i}|\mathbf{x}_i) = P(a = \hat{y}^c_{i} | s = \mathbf{x}_i)$ for each action condition on the current state. We adopt the model structure commonly utilized in previous supervised studies as the architecture for our policy network. 

\paragraph{Rewards} 

Recall that our primary objective is to acquire a less biased label distribution compared to the supervised negative mode training approach using the partially annotated training dataset. We anticipate that our MLPAC possesses the capacity for balanced consideration of both \textit{exploitation} and \textit{exploration}. \textit{Exploitation} ensures that our agent avoids straying from local optima direction and avoids engaging in excessively invalid exploratory behavior, while \textit{exploration} motivates our agent to explore the action space somewhat randomly and adapt its policy, preventing overfitting to partial supervision.
Inspired by the actor-critic RL algorithm \citep{bahdanau2016actor}, we design our rewards function containing two parts: a \textbf{local reward} provided by a trainable critic network, which provides immediate value estimation of each action and a \textbf{global reward} regarding the overall performance of all classes predictions for each instance. 

Specifically, inspired by \citet{luo2021pulns}, the local reward calculates the reward of each class prediction for each instance according to the critic network confidence:
\begin{align}
        &r_{i}^{c}  (V_\lambda,  \mathbf{x}_i, c)\,    \\
        &=\,\left\{\begin{array}{l l} {{\mathbb{C}(-1,\,\mathrm{log}\,\frac{p_{V_\lambda}^c(\mathbf{x}_i)}{1-p_{V_\lambda}^c(\mathbf{x}_i)},\,1)}}&{{\mathrm{if}\;\hat{y}^c_{i}=1,}}\\
        {{\mathbb{C}(-1,\,\mathrm{log}\,\frac{1-p_{V_\lambda}^c(\mathbf{x}_i)}{p_{V_\lambda}^c(\mathbf{x}_i)},\,1)}}&{{\mathrm{if}\;\hat{y}^c_{i}=-1.}}
        \end{array}\right. \nonumber
\label{eq:local_reward}
\end{align}   
where $p_{V_\lambda}^c(\mathbf{x}_i)$ and $1-p_{V_\lambda}^c(\mathbf{x}_i)$ are the probabilities of class $c$ being \texttt{TRUE} and \texttt{FALSE} respectively for an instance $\mathbf{x}_i$, calculated by a critic network $V_{\lambda}$ with parameter $\lambda$, 
$\hat{y}^c_{i}$ denotes the prediction of policy network, 
and $\mathbb{C}(-1, \cdot, 1)$ is a clamping function:
a) $\mathbb{C}(-1, x, 1) = -1$ if $x < -1$; b) $\mathbb{C}(-1, x, 1) = 1$ if $x > 1$; c) otherwise, $\mathbb{C}(-1, x, 1) = x$. 

We train the critic network in a supervised fashion (under negative mode) to guide the policy network's exploration direction through the local reward function, thus equipping our framework with the ability of \textit{exploitation}. 
From Equation \ref{eq:local_reward}, the action of our policy network could get a positive reward if the critic network has the same prediction tendency and a negative reward otherwise.  
The policy network is rewarded to ``softly fit'' the distribution learned by the critic network. To improve the accuracy of value estimation, we perform label enhancement to train our critic network (described in \cref{sec:learn}).
Consequently, the local rewards offer a preliminary yet instructive signal to guide the learning process in our MLPAC framework, thereby
preventing the MLPAC from engaging in excessively invalid exploratory behavior within the large action space, thereby enhancing the overall learning efficiency. Nevertheless, relying solely on these local rewards may potentially lead the MLPAC system to converge to a biased ``negative mode'' solution. To mitigate this risk, we introduce global rewards to stimulate more comprehensive exploration during the learning process.

As for \textit{global} reward, we employ a straightforward yet highly effective scoring function, which is computed based on the recall metric. In detail, for the whole classes prediction $\hat{\mathbf{y}}_i$ of $\mathbf{x}_i$ with the observed ground truth $\mathbf{y}_i$, the recall score is:
\begin{align}
    &recall(\mathbf{y}_i, \hat{\mathbf{y}}_i) \\
    &= \frac{|\{y_i^c=1 \wedge \hat{y}^c_i = 1, \,\, y_i^c \in\mathbf{y}_i, \hat{y}^c_i \in \hat{\mathbf{y}}_i\}|}{|\{y^c_i = 1, \,\, y^c_i \in \mathbf{y}_i\}|} \nonumber
\end{align}
To enhance recall scores, our policy network is encouraged to predict a greater number of classes as \texttt{TRUE}, thereby alleviating the catastrophic label imbalanced challenge \footnote{Note that we do not punish the action of wrongly setting a class as \texttt{TRUE}. Thus, the policy network is pleased to predict all classes as \texttt{TRUE} if we only leverage the global reward function}.

Note that in our reward design, the terms ``local'' and ``global'' are both used to characterize the ``goodness'' of the predictions of input instances (i.e., state-action pairs). They are both immediate rewards in the considered RL framework, as we formalize MLPUL as a one-step MDP. To calculate the final reward of the whole predictions of an instance, the local rewards of all predicted classes $c$ are summed out, eventually weighted summed with global reward:
\begin{align}
\label{eq:total_reward}
    &R(\mathbf{x}_i, \hat{\mathbf{y}}_i, V_\lambda, \mathbf{y}_i) 
    \\ \nonumber
    & = \frac{1}{|\mathcal{C}|}\sum_{c\in \mathcal{C}} r_i^c(V_\lambda, \mathbf{x}_i, c) + w * recall(\mathbf{y}_i, \hat{\mathbf{y}}_i), 
\end{align}
where $w$ is a weight controlling the scale balance between local reward and global reward.

\subsection{Inference}
The final predictions of each instance are decided according to the probabilities that our policy network output. We simply set classes whose probabilities are more than 0.5 ($\pi_\theta(\hat{y}^c_{i}|\mathbf{x}_i) > 0.5$) to as \texttt{TRUE} (\textit{i.e.}, $\hat{y}^c_{i}=1$).

\subsection{Learning}
\label{sec:learn}
We iteratively train our critic network and policy network in an end-to-end fashion. Since the critic network plays a critical role in guiding policy network learning, we employ label enhancement techniques during the training of the critic network to enhance the precision of value estimations. 
It is important to emphasize that we intentionally exclude the enhanced labels from participation in the calculation of the recall reward. This decision is motivated by the desire to maintain the precision of the global reward and prevent potential noise introduced by the enhanced labels. 

It is widely acknowledged that the training process in RL can experience slow convergence when confronted with a vast exploration space. Inspired by previous RL-related works \citep{silver2016mastering, qin2018robust}, we initiate our process by conducting pre-training for both our policy and critic networks before proceeding with the RL phase. 
Typically, pre-training is executed through a supervised method. In our settings, a range of trivial solutions for MLPUL can serve as suitable candidates for pre-training. In most cases, we simply utilize the negative mode for pre-training. However, as previously mentioned, the negative mode tends to acquire a biased label distribution due to severe label imbalance. 
Thus, we implement an early-stopping strategy during the pre-training phase to prevent convergence.
The following introduces the detailed learning strategies and objectives.

\begin{algorithm}[h!]
    \DontPrintSemicolon
    \SetNoFillComment
    \caption{Partially Annotated Policy Gradient Algorithm}
    \label{alg:MLPACa}
        \KwIn {Observed data $\mathcal{X}$, partial labels 
        $\mathcal{Y}$, pre-trained policy network $\pi_{\theta^0}$, critic network $V_{\lambda}$, REINFORCE learning rate $\alpha$, confidence threshold $\gamma$, sample steps $T$ }
        \KwOut {Optimal parameters $\theta^*$}
        $e \gets 0$, $\theta^* \gets \theta^0$, enhanced annotation set $\bar{\mathcal{Y}}  \gets \mathcal{Y}$\;
        \While{$e < $  total training epoches}{
        Training set for critic network: $(\mathcal{X}, \bar{\mathcal{Y}})$\;
        Training set for policy network: $(\mathcal{X}, \mathcal{Y})$\; 
        \For{ $\mathcal{X}_{batch} \in \mathcal{X}$ in total batches }{
        Update $\lambda$ by minimizing Equation~\ref{eq:loss_sup} with 
        $\bar{\mathcal{Y}}_{batch}$
        for critic network

        \For{{\rm step} $t < $ {\rm sample steps} $T$ }{
                For each $\mathbf{x}_i \in \mathcal{X}_{batch}$, sample $\hat{\mathbf{y}}_i$ w.r.t. $\hat{\mathbf{y}}_i \sim \pi_\theta (\hat{\mathbf{y}}_i | \mathbf{x}_i)$\;
                Compute $R(\mathbf{x}_i, \hat{\mathbf{y}}_i, V_\lambda, \mathbf{y}_i)$ according to Equation~\ref{eq:total_reward}\;
            }
            Update policy network using $\theta \gets \theta + \alpha \nabla_\theta \mathcal{J}_{PG}(\theta)$\;
        }
        $\bar{\mathcal{Y}} \gets \{ [\bar{y}_i^c]_{c=1}^{|\mathcal{C}|}\}$, where $\bar{y}_i^c$ = 1 if ($y_i^c=1$ or \big($\pi_\theta (\hat{y}_i^c =1| \mathbf{x}_i)>\gamma \,\, \text{and} \,\, \hat{y}_i^c = V_\lambda(\mathbf{x}_i)_c = 1$  )\big)  \;
        \If{eval$(\pi_\theta)$\,\, $>$ \,\, eval$(\pi_{\theta^{*}})$}{
            $\theta^* \gets \theta$\;
        }
        $e \gets e + 1$\;
        }

        \Return{$\theta^*$}\;
\end{algorithm}

\paragraph{Objective for Value Model}
Generally, a well-designed supervised objective urges models to learn expected outputs by learning from annotated data. This process typically refers to the \textit{exploitation}, where the supervised model fits the distribution of label annotations. We denote the supervised objective by a general formulation:
\begin{equation}
    \mathcal{L}_{SUP}(\theta) = \sum\nolimits_{\mathbf{x}_i \in \mathcal{X}} p(\mathbf{x}_i) \mathcal{D}(\mathbf{y}_i, \hat{\mathbf{y}}_i),
    \label{eq:loss_sup}
\end{equation}
where $\mathcal{D}$ is a task-specific distance metric measuring the distance between annotation $\mathbf{y}_i$ and prediction $\hat{\mathbf{y}}_i$. 
Recall that we treat all the unknown classes as negatives to perform supervised learning.

\paragraph{Objective for Policy Model}
As stated in previous work \citep{qin2018robust}, policy-based RL is more effective than value-based RL in classification tasks because the stochastic policies of the policy network are capable of preventing the agent from getting stuck in an intermediate state.
We leverage policy-based optimization for RL training. The objective is to maximize the expected reward:
\begin{align}
    &\mathcal{J}_{PG}(\theta) = \mathbb{E}_{\pi_\theta} [R(\theta)] \\
    &\approx \sum_{\mathbf{x}_i \in batch} p(\mathbf{x}_i)\sum_{\hat{\mathbf{y}_i} \sim \pi_\theta (\hat{\mathbf{y}_i} | \mathbf{x}_i)} \pi_\theta (\hat{\mathbf{y}_i} | \mathbf{x}_i) R(\hat{\mathbf{y}_i}. \mathbf{x}_i), \nonumber
\end{align}
The policy network $\pi_\theta$ can be optimized w.r.t. the policy gradient REINFORCE algorithm~\citep{williams1992simple}, where the gradient is computed by
\begin{align}
    &\nabla_\theta \mathcal{J}_{PG}(\theta) = \sum_{\mathbf{x}_i \in batch} p(\mathbf{x}_i)\\
    &\sum_{\hat{\mathbf{y}_i}  \sim \pi_\theta (\hat{\mathbf{y}_i} | \mathbf{x}_i)} \nabla_\theta \ln (\pi_\theta (\hat{\mathbf{y}_i} | \mathbf{x}_i)) R(\hat{\mathbf{y}_i}, \mathbf{x}_i), \nonumber
    \label{eq:loss_pg}
\end{align}
where $p(\mathbf{x}_i)$ is a prior distribution of input data. Specific to uniform distribution, $p(\mathbf{x}) = \frac{1}{|\mathbf{x}_{batch}|}$.

\paragraph{Overall Training Procedure}
The overall training process is demonstrated in Algorithm \ref{alg:MLPACa}, where $\hat{y}_i^c = V_\lambda(\mathbf{x}_i)_c = 1$ refers to the prediction of class $c$ that critic network outputs for the sample $\mathbf{x}_i$ being $\texttt{TRUE}$ ($p_{V_\lambda}^c(\mathbf{x}_i)>0.5$). 
There are several strategies that need to be clarified in the overall training procedure. We empirically prove the effectiveness of these strategies.

\noindent $\rhd$ \quad  The Computation of local rewards is based on the annotated positive classes and a randomly selected subset of unknown classes rather than considering the entire class set of an instance, which is intended to emphasize the impact of positive classes within the computation of local rewards. 

\noindent  $\rhd$  \quad The enhanced labels are determined by both the policy network and critic network to guarantee high confidence of enhanced positive labels.


\noindent  $\rhd$ \quad  We fix the critic network once it converges to enhance training efficiency.



\begin{table*}[t!]

\setlength\tabcolsep{3pt}
\small
\centering
\resizebox{0.9\textwidth}{!}{%

\begin{tabular}{l|lll|lll|lll|lll|lll}
\toprule
                & \multicolumn{3}{c|}{\textbf{10\%} }                                                     & \multicolumn{3}{c|}{\textbf{30\%} }     & \multicolumn{3}{c|}{\textbf{50\%} }
                & \multicolumn{3}{c|}{\textbf{70\%} }                                       & \multicolumn{3}{c}{\textbf{100\%}}                                       \\ 
     {\textbf{Method}}   & {P} & {R} & {F1} & {P} & {R} & {F1} & {P} & {R} & {F1} & {P} & {R} & {F1}& {P} & {R} & {F1}\\ \midrule

\textbf{SSR-PU (fix-prior)} & 75.5 & 35.8 & 48.6 & 52.4 & 69.9 & 59.9 & 33.6 & 83.0 & 47.8 & 23.3 & 87.2 & 36.8 & - & - & - \\

\textbf{SSR-PU (vary-prior)}  & 79.6 & 33.1 & 46.7 & 82.1 & 56.1 & 66.7 & 83.1 & 67.0 & 74.2 & 83.0 & 71.6 & 76.9 & - & - & 78.9 \\

\midrule

\textbf{Negative Mode} & 89.8  &  3.8 & 7.3 & 92.0 & 20.8 & 33.8 & 91.8 & 42.1 & 57.7 & 89.6 & 58.8 & 70.6 & 86.0 & 72.4 & 78.6
\\

\textbf{Pos Weight} &  84.9  & 39.8  & 54.1 & 85.5 & 59.3 & 70.0 & 85.0 & 66.8 & 74.8 & 83.4 & 72.5 & 77.6 & 82.9 & 77.3 & 80.0
\\ 
\textbf{Neg Weight} &  86.7  & 30.7  & 45.4 & 85.0 & 59.3 & 69.8 & 84.1 & 68.2 & 75.3 & 82.8 & 72.8 & 77.5 & 79.8 & 78.7 & 79.3
\\
\textbf{MLPAC~(Ours)} & 58.5  & 77.0  & 66.0 & 83.5 & 67.71 & 74.7 & 81.4 & 73.6 & 77.3 & 83.3 & 73.9 & 78.3 & 80.9 & 80.8 & 80.9
\\
\midrule
\midrule
\textbf{P3M (fix-prior)}  & 81.8 & 50.6 & 62.5 & 74.3 & 76.4 & 75.3 & 67.6 & 84.0 & 75.0 & 62.0 & 87.6 & 72.6 & - & - & -  \\

\textbf{P3M (vary-prior)} & 76.6 & 59.2 & 66.8 & 73.2 & 77.2 & 75.1 & 70.5 & 82.4 & 76.0 & 69.4 & 84.3 & 76.2 & - & - & 80.0 \\
\bottomrule

\end{tabular}
}
\caption{Results on Re-DocRED with varying ratios of positive class annotations. \textbf{Fix-prior} means that we keep the same ``true positives/observed positives'' prior  (=3) in their methods under different ratios, while \textbf{vary-prior} means that we set the actual prior corresponding to the rations. The concurrent method \textbf{P3M} is shown for reference.}
\label{tab:re}
\end{table*}

\section{Experiments}\label{sec:exp}

The Multi-Label Positive Unlabelled Learning (MLPUL) problem is common and essential in the NLP field. There are many tasks in NLP that face incomplete annotation problems, such as fine-grained entity typing, multi-label text classification, and document-level relation extraction.
To verify the effectiveness of our proposed RL framework, we experiment with the positive-unlabelled document-level relation extraction (DocRE) task as a representative MLPUL task in NLP. We also conduct experiments in multi-label image classification (MLIC) tasks and \textit{binary} PU learning setting to verify the generalization and effectiveness of our framework.

Besides comparing with previous state-of-the-art (SOTA) models in each specific task, we construct some simple baseline methods:
\begin{itemize}[leftmargin=*, noitemsep, topsep=0pt]
    \item \textbf{Negative Mode}: As mentioned in \cref{sec:problem_set}, all unknown labels are treated as negative labels, performing conditional supervised learning.
    \item \textbf{Pos Weight}: Based on negative mode, we up-weight positive labels' loss in the supervised loss.
    \item \textbf{Neg Weight}: Based on negative mode, we perform negative sampling in the supervised loss.
\end{itemize}

We train the model on 1 NVIDIA A100 GPU. The total number of training epochs is 30. We iteratively train our critic and policy network for the first 10 epochs, and then we only train the policy network for the last 20 epochs. 
The threshold $\gamma$ of choosing enhancement labels is $0.95$ in most cases, and we find our framework performs robustly to $\gamma$ varying from $0.5$ to $0.95$.
We tune the hyper-parameter reward weight $w$ in Eq.\ref{eq:total_reward} and sampling number $T$ with different experiment settings, 
and $w$ is dynamically adjusted during training\footnote{Intuitively, the $w$ of recall reward should be dropped along with the training epochs because our critic network provides more and more accurate local rewards beneficial by data enhancement before convergence.}.
All the above hyper-parameters are determined according to validation set performance.
We leverage the F1 score and MAP score (for MLIC) to evaluate all the models.
We conduct experiments on selected datasets with varying ratios of positive class annotations from 10\% to 100\%. 
We randomly keep a ratio of annotated relations and treat all the leaving classes as unknown.
Part of the results are shown in the experimental tables.
Detailed descriptions of these simple baselines, evaluation metrics, data statistics, and full experimental results can be found in Appendix \ref{appd:tech_details}, \ref{appd:data}, \ref{appd:config}, and \ref{appd:more_exp} respectively.





\subsection{Document-level Relation Extraction}\label{sec:doc_exp}

Document-level Relation Extraction (DocRE) is a task that focuses on extracting fine-grained relations between entity pairs within a lengthy context. Align to our formulation, an input $\mathbf{x}_i$ is an entity pair, and $\mathbf{y}_i$ represents relation types between the entity pair. An entity pair may have multiple relations or have no relation in DocRE.
Beyond the experiments trained on an incomplete annotated DocRED training set and tested on an almost fully annotated Re-DocRED test set (as shown in \cref{fig:intro}C), we choose the Re-DocRED, which is the most complete annotated dataset in DocRE, for experiments on varying ratio of positive annotations. The size of class set $\mathcal{C}$ is $97$ (contains \texttt{<None>}). 

\paragraph{Configuration and Baselines}
We adopt the fully supervised SOTA, DREEAM \citep{ma2023dreeam}, as our critic and policy network architectures in this experiment. We keep the same training method with an Adaptive Thresholding Loss (ATL) of DREEAM for our critic network.
The hyper-parameter $w$ of reward weight in \cref{eq:total_reward} is set to 10. The sample steps $T$ in RL is set to 10. We compare our method to \textbf{SSR-PU} \cite{wang2022unified} and show the results of concurrent work \textbf{PM3} \cite{wang2023positive} for reference. Both of these two models perform rank-based PU loss with an assumption of label distribution prior.

\paragraph{Results}
Experimental results in Re-DocRED are shown in Table~\ref{tab:re}. Compared to previous work and our simple baselines, our \textbf{MLPAC} demonstrates its advantage in all annotation ratios. Besides, our framework achieves more balanced precision-recall scores, suggesting its ability to deal with imbalanced label challenges and predict more positive labels. 
It is worth noting that our framework also achieves improved performance with the full annotated dataset because the full annotations of Re-DocRED still miss some actual relations, as mentioned in \citet{li2023semi}. Furthermore, our framework does not rely on label prior estimation, while previous rank-based methods are sensitive to a certain extent (fix-prior vs. vary-prior).

\subsection{Multi-Label Image Classification}\label{sec:image_exp}

Multi-label image classification is the task of predicting a set of labels corresponding to objects, attributes, or other entities present in an image.
Following previous work \cite{kanehira2016multi, ben2022multi},
we utilize MS-COCO dataset \citep{lin2014microsoft} containing 80 classes. 


\paragraph{Configuration and Baselines}
Our policy and critic networks adopt the 
same architecture as P-ASL \cite{ben2022multi}. We rerun previous works (\textbf{ERP}, \textbf{ROLE}, \textbf{P-ASL}) with official code \cite{cole2021multi,ben2022multi} for fair comparisons with our methods under the same training data samples. We tune the hyper-parameter $w$ between \{5, 7, 12\} in this task. The sample steps $T$ in RL is set to 50. Detailed descriptions of compared models can be found in Appendix \ref{appd:config}. 
\paragraph{Results}
Experimental results are shown in \cref{tab:coco}. 
Our \textbf{MLPAC} still performs competitively in the MLIC task. 
Furthermore, we ran our model three times and found very small standard deviations of F1 scores and mAP, which demonstrates the high robustness and stability of our framework.
Standard deviations of three runs and more experimental results can be found in Appendix \ref{appd:img_exp}.


\begin{table}[t!]
\small
\centering

\label{tab:coco}
\setlength{\tabcolsep}{3pt}{
\begin{tabular}{l|cc|cc|cc}
\toprule
               & 
               \multicolumn{2}{c|}{\textbf{30\%}}&
               \multicolumn{2}{c|}{\textbf{50\%}}&
               \multicolumn{2}{c}{\textbf{70\%}}
              
               \\ 
{\textbf{Method}}   & {F1} & {mAP} & {F1} & {mAP} & {F1} & {mAP} \\ 
     \midrule
\textbf{ERP}    & - & 71.0 & - & 73.5 & - & 73.8 \\
\textbf{ROLE}   & - & 72.4 & - & 76.6 & - & 79.5 \\
\textbf{P-ASL+Negative} & 52.1 & 74.6 & 54.0 & 76.9 & 71.9 & 81.0 \\
\textbf{P-ASL+Counting}  & 26.4 & 63.4 & 53.7 & 76.1 & 71.6 & 80.1 \\
\midrule
\textbf{Negative Mode} & 33.7 & 64.3 & 52.9 & 73.8 & 72.3 & 81.2
\\ 
\textbf{Pos Weight} & 73.0 & 72.7 & 75.7 & 76.7 & 76.0 & 79.9
\\ 
\textbf{Neg Weight}  & 68.7 & 74.8 & 75.9 & 78.0 & 77.9 & 79.7
\\

\textbf{MLPAC~(Ours)} & 77.0 & 77.5 & 79.1 & 80.4 & 79.0 & 81.4 \\
\bottomrule
\end{tabular}
}
\caption{Experimental Results on COCO datasets with varying ratios of positive classes annotations.}
\end{table}

\begin{figure}[t]
\small
    \includegraphics[width=0.9\linewidth]{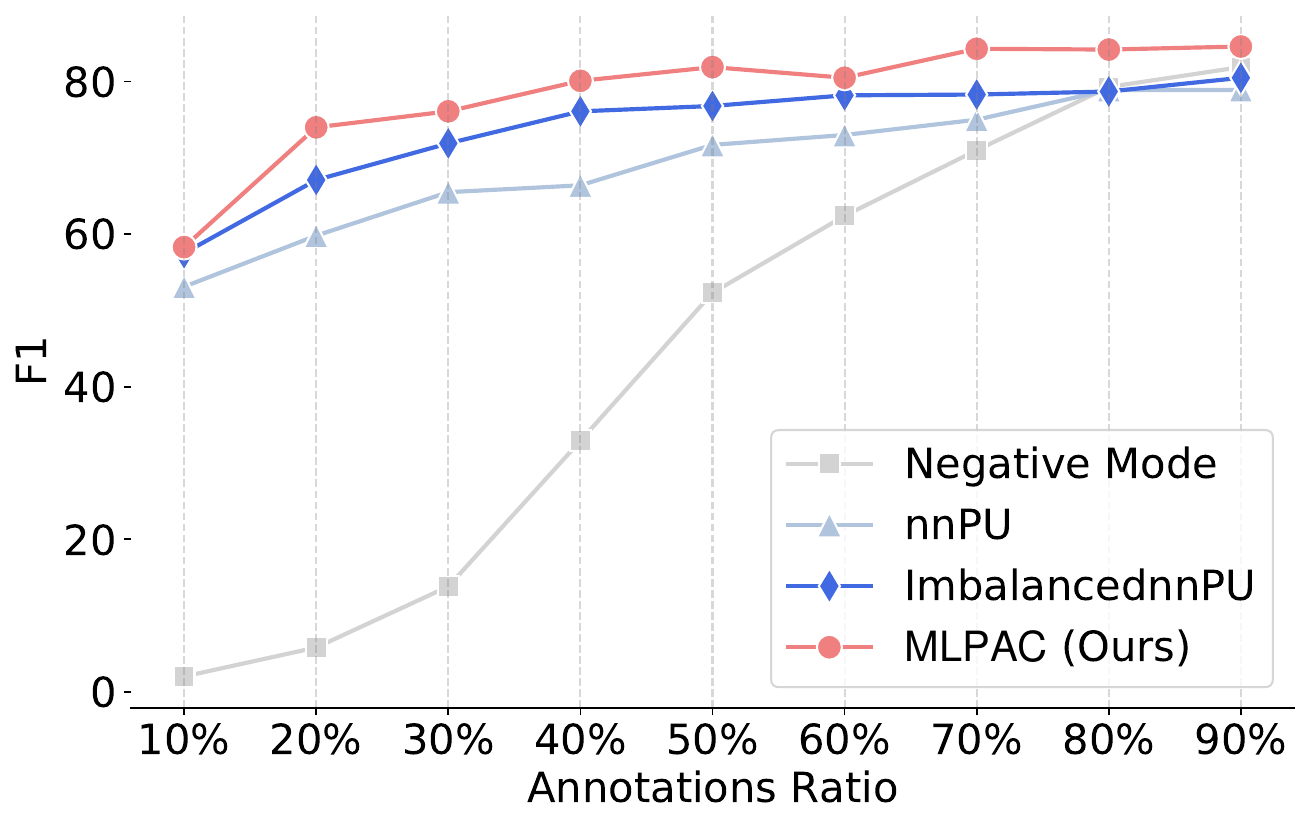}
    \caption{Experimental results of the setting with ``truck'' category as positives.}
    \label{fig:toy1}
\end{figure}

\begin{table}[t!]
\small
\centering
\setlength{\tabcolsep}{3pt}{
\begin{tabular}{l|lll|lll}
\toprule
                & \multicolumn{3}{c|}{\textbf{Re-DocRED} }                                                                                           & \multicolumn{3}{c}{\textbf{COCO}}                                       \\ 
     {\textbf{Method}}   & {P} & {R} & {F1} & {P} & {R} & {F1} \\ \midrule
\textbf{MLPAC~(Ours)}   & 64.5  & 72.8 & 68.4  &80.9 & 59.2 & 68.3   \\
\textbf{ w/o. Local reward}     & 12.2 & 93.3  & 21.6 & 61.5 & 65.5  & 63.4    \\
\textbf{ w/o. Global reward}     & 84.5 &  45.9 & 59.5 & 89.7 & 6.9& 12.8\\
\textbf{w. Prec }   & 85.9  & 44.0 & 58.1  & 96.2 & 29.8 &  45.5 \\
\textbf{w. F1 }   & 86.0  & 43.3 & 57.6  & 89.2 & 46.8 & 61.4 \\

\bottomrule
\end{tabular}
}
\caption{Ablation study on our rewards.}
\label{tab:reward}
\end{table}

\begin{table}[t!]
\small
\centering
\setlength{\tabcolsep}{3pt}{
\begin{tabular}{l|lll|lll}
\toprule
                & \multicolumn{3}{c|}{\textbf{Re-DocRED} }                                                                                           & \multicolumn{3}{c}{\textbf{COCO}}                                       \\ 
     {\textbf{Method}}   & {P} & {R} & {F1} & {P} & {R} & {F1} \\ \midrule
\textbf{MLPAC~(Ours)}     & 64.5 &  72.8 & 68.4 & 80.9 & 59.2 & 68.3   \\
\textbf{ w/o. Iterative training}     & 89.9 &  34.6 & 49.9 & 76.4 & 33.8 & 46.9   \\
\textbf{ w/o. Label enhancement}     & 83.6 & 47.2  & 60.3 & 51.7 & 54.8 & 53.2  \\
\textbf{ w/o. Action sampling}     & 88.2 &  36.5 & 51.7 & 96.4 & 20.4 & 33.7  \\
\textbf{ Supervised self-training} &  68.0  &  29.0 & 40.7 & 89.7 & 6.6 &  12.3
\\ 
\bottomrule
\end{tabular}
}
\caption{Ablation study on our training strategy.}
\label{tab:training}
\end{table}

\subsection{Binary PU Learning Setting}\label{sec:toy_exp}

To verify the generalization and wide adaptation of our RL framework, 
we conduct binary PU learning with the same setting following \citet{su2021positive} that concerns positive/negative imbalanced problems in binary image classification. 
The imbalanced datasets are constructed from CIFAR10\footnote{A multi-class dataset containing ten categories. \url{https://web.cs.toronto.edu/}} by picking only one category as positives and treating all other data as negatives. 


\paragraph{Configuration and Baselines} 
Of note, our framework can integrate any supervised model architecture. For a fair comparison, we take the same architecture of \citet{kiryo2017positive, su2021positive} as our critic and policy networks. 
We compared to \textbf{nnPU} \cite{kiryo2017positive} and \textbf{ImbalancednnPU} \cite{su2021positive}.
More details are in Appendix \ref{appd:config}.
We tune the hyper-parameter $w$ between $\{10, 20, 50\}$.
The action sampling number $T$ is $100$.


\paragraph{Results} We show F1 scores with varying ratios of annotated positives in \cref{fig:toy1}.
Our \textbf{ MLPAC} achieves significant improvements over \textbf{Negative Mode} and previous work, 
It is worth mentioning that our framework still demonstrates its superiority though \textbf{ImbalancednpU} is specifically designed for binary PU learning with imbalanced settings. 

\subsection{Analysis}
We conduct ablation studies to analyze our MLPAC framework both on modeling and training strategy.  

\paragraph{Rewards Design:}
To show the effectiveness of combining \textit{exploitation} and \textit{exploration} and the benefit of \textit{local} and \textit{global} rewards, we train our framework in the $10\%$ annotations setting without local rewards and global rewards, respectively. Additionally, we replace the recall scores with precision \textbf{w. Prec} or F1 scores \textbf{w. F1} as our global rewards to show the effects of different global reward designs. Experimental results are shown in Table \ref{tab:reward}. It can be observed that it is hard for an RL framework to achieve comparable performance without local rewards to guide exploitation. The reason is that the action space of multi-label classification is too large to find the global optimal directions. Without our global reward, the recall evaluation score drops a lot (72.78 vs. 45.94), which demonstrates the advantage of the global reward in alleviating imbalance distribution. Both the two variants of global reward damage the performance, revealing the advance of taking the exactly accurate evaluation as rewards in the partially annotated setting.


\paragraph{Training Strategy:}

To verify the effectiveness of our training procedure, we attempt different training strategies shown in Tabel \ref{tab:training}. 
\textbf{w/o. Iterative training} means that we fix the critic network after pretraining and only train the policy network in the RL training procedure. 
\textbf{w/o. Data enhancement} means that we still iteratively train our critic and policy network but do not enhance pseudo labels for the critic network. 
\textbf{w/o. Action sampling} means that we leverage the whole action sequence to calculate local rewards without sampling operation illustrated in \cref{sec:learn}. 
\textbf{Supervised self-training} means that we conduct self-training of the critic network. 
It is obvious that our training method makes remarkable achievements. 
More analysis experiments are in Appendix \ref{appd:doc_exp}.








\section{Conclusion}

In this work, we propose an RL framework to deal with partially annotated multi-label classification tasks. We design local rewards assessed by a critic network and global rewards assessed by recall functions to guide the learning of our policy network, achieving both exploitation and exploration.
With an iterative training procedure and a cautious data enhancement, our MLPAC has demonstrated its effectiveness and superiority on different tasks.  

\section*{Limitation}
We have considered the label correlations in our challenge. However, in our setting, label correlations may bring correlation bias to the severe scarcity of positive labels. Therefore, we did not explicitly model the correlations in our current framework. We would like to explore the potential of leveraging the label correlations to enhance our framework in future work.

\section*{Acknowledgements}
The authors thank the reviewers for their insightful suggestions on improving the manuscript. This work presented herein is supported by the National Natural Science Foundation of China (62376031).

\bibliography{anthology,custom}

\appendix


\section{Related Work}
\label{appd:related}
\paragraph{Different Settings under Weak Supervision}
Many previous works, such as partial conditional random field, focus on single-label multi-class tasks with partial supervision \citep{mayhew2019named, effland-collins-2021-partially, li2021empirical, zhou-etal-2022-distantly}. 
Recently, there have been various settings of multi-label classification tasks without fully annotated training sets:
\underline{\textit{Multi-label positive unlabelled learning}}, which is our concern, supposing a multilabel dataset has properties by which (1) assigned labels are definitely positive and (2) some labels are absent but are still considered positive \cite{kanehira2016multi, wang2022unified, yuan2023positive};
\underline{\textit{Partially observed labeling}}, which is also termed as \textit{Multi-label classification with missing labels} supposes partial positive \textit{and} negative classes are labeled \cite{durand2019learning, zhang2022effective, ben2022multi, abdelfattah2022g2netpl};
\underline{\textit{Partially positive observed labeling}}, which supposes only part of positive classes are observed \textit{with the assumption} that at least one positive label per instance should be observed.
\underline{\textit{Single positive labeling}}, which supposes one and \textit{only} one positive class per instance is observed \cite{cole2021multi, kim2022large, jouanneau2023patch}. 
\underline{\textit{Distantly supervised learning}}, which supposes annotated samples contain both false positives and false negatives, devoted to dealing with label noise problems \cite{ye2020deep, sun2023uncertainty, zeng2024document}.


\section{More technology details}
\label{appd:tech_details}

\subsection{Pos Weight and Neg Weight}
\label{appd:pos_neg}
In the ``Pos Weight'' method, we impose a large weight $w_p$ to the positives. We set $w_p$ as the times of spositive targets to unlabeled targets in each training batch.
Previous study \citep{li2020empirical} had stated that negative sampling can be considered as a type of negative weighting method. And this work experimentally find that negative sampling even work better. In our experiments, we under-sampling the unlabeled targets as the ``Neg Weight'' method. Unlabeled targets 10 times the number of positive targets are retained in each training batch.

\subsection{Evaluation Metrics}
\label{appd:eva_met}
We compute the F1 scores based on TP (True Positive), FP (False Positive), and FN (False Negative).
\begin{equation}
  \begin{aligned}
      &Recall = TP / (TP + FN),\\
      &Precision = TP / (TP + FP),\\
      &F1 = \frac{2*Recall*Precision}{Recall + Precision}.
  \end{aligned}
\end{equation}

We choose the widely used evaluation metric mAP on multi-label image classification. $N_c$ is the number of images containing class $c$, Precision($k$, $c$) is the precision for class $c$ when retrieving $k$ best predictions and rel($k$, $c$) is the relevance indicator function that is 1 if the class $c$ is in the ground-truth of the image at rank $k$. We also compute the performance across all classes using mean average precision (mAP), where $C$ is the number of classes.
\begin{align}
    &AP_c = \frac{1}{N_c} \sum_{k=1}^{N} Precision(k, c) * rel(k, c), \\ 
      & mAP = \frac{1}{C} \sum_{c} AP_c
\end{align}

\section{Data Statistics}
\label{appd:data}
\paragraph{Document-level Relation Extraction}

Given a document $\mathbf{x}$ containing entities $\mathcal{E}_\mathbf{x} = \{e_i\}^{|\mathcal{E}_\mathbf{x}|}_{i=1}$, DocRE aims to predict all possible relations between every entity pair. Each entity $ e \in \mathcal{E}_\mathbf{x}$ is mentioned at least once in $\mathbf{x}$. with all its proper-noun mentions denoted as $\mathcal{M}_e = \{m_i\}^{|\mathcal{M}_e|}_{i=1}$. Each entity pair $(e_s, e_o)$ can hold multiple relations, comprising a set $\mathbf{y} = \mathcal{R}_{s,o} \subset \mathcal{R}$, where $\mathcal{R}$ is a pre-defined relation set. We let the set $\mathcal{R}$ include $\epsilon$, which stands for \textit{no-relation}. To better formulate, we denote the target of DocRE for each document as a set of multi-hot vectors representing labels of relation-existing entity pair $\{\mathbf{y}(e_s, e_o) = \mathbf{y}^{so} = [y^{so}_1, ..., y^{so}_i, ..., y^{so}_{|\mathcal{R}|}], s\in\{1,|\mathcal{E}_D|\}, o\in\{1,|\mathcal{E}_D|\}\}$, where $y^{so}_i\in\{0,1\}$ and $\sum_{i=1}^{|\mathcal{R}|}y_i=|\mathcal{R}_{s,o}|$.

In this task, we have trained our model based on the training set in the Re-DocRED dataset and validated our model by the Re-DocRED test set and DocGNRE test set. There are 3053 documents (including 59359 entities, and 85932 relations) in the Re-DocRED training set and 500 documents (including 9779 entities, and 17448 relations) in the Re-DocRED test set. DocGNRE test set provides a more accurate and complete test set with the addition of 2078 triples than ReDocRED.

To simulate partial annotation, we randomly kept a ratio of annotated relations.
As mentioned in the introduction, Re-DocRED still misses some actual relation annotations. Hence, we also conduct experiments on the full training set to compare our framework with previous fully-supervised work. The size of label set $\mathcal{C}$ is $97$ (contains \texttt{<None>}) in this task.  
Supervised learning on DocRE generally faces two challenges: i) lots of actual relation triples are not labeled by annotators because of the long context and fine-grained relation types; ii) the number of \textit{no-relation} entity pairs are far larger than the number of relation-existing entity pairs, which cause an unbalanced problem.

\paragraph{Multi-label Image Classification}

Following previous work \cite{kanehira2016multi, ben2022multi}, \citet{ben2022multi} (\textbf{P-ASL}), which deals with partial annotations containing both positive and negative classes, 
we utilize MS-COCO dataset \citep{lin2014microsoft} containing 80 classes. We keep the original split with 82081 training samples and 40137 test samples. We simulate the partial annotations following the operations in \textbf{P-ASL}. But different from them, we only retain the positive classes in their partial annotations and take all the rest of the classes as \texttt{UNKNOWN}. Specifically, We utilize the MS-COCO dataset to simulate the `Random per annotation' scheme. We omit each annotation no matter the positive or negative label with probability $p$. With our setting, the retained positive labels are considered as positives and the rest are all unlabeled. We will not directly exploit unavailable negative annotations.

\paragraph{Binary PU Learning Setting}
With our formulation, an instance $\mathbf{x}_i$ is an image, and the label of an instance in binary classification settings can be denoted as $\mathbf{y}_i = [y_i^1, y_i^2]$ where $y_i^1$ is the label of positive and $y_i^2$ is the label of negative. The prediction for each image is conducted by setting $y$ corresponding to the higher score as $1$ and the other as $0$. Hence, there are 50,000 training data and 10,000 test data as provided by the original CIFAR10. To make the training data into a partially annotated learning problem, we randomly sample a ratio of positives as annotated data and all the leaving training data as an unknown set.

\section{Detailed Experimental Configuration}
\label{appd:config}
\paragraph{Document-level Relation Extraction}

We randomly sample three different versions of datasets and report the average results over them. Detailed scores are in \cref{tab:ns_com} in Appendix \ref{appd:doc_exp}. The training hyper-parameters are the same as DREEAM \citep{ma2023dreeam}. We only train the policy network for the last 20 epochs. It takes about 6 hours.

\paragraph{Multi-Label Image Classification} 

For a fair comparison, our critic and policy networks have the same architecture as \textbf{P-ASL}. The training hyper-parameters are the same as that in \cite{ben2022multi}. Due to the different partially annotated settings, we rerun \textbf{P-ASL} utilizing their codebase but with our datasets. 
\textbf{P-ASL+Negative} means training a model taking all \texttt{UNKNOWN} as negative classes to predict label distribution as prior. \textbf{P-ASL+Counting} means counting partially labeled positive classes as distribution prior. We also rerun \textbf{EPR} and \textbf{ROLE} methods from \cite{cole2021multi} with our datasets, utilizing their official code.
We tune the hyper-parameter $w$ between $\{5, 7, 12\}$ in this task. Following previous work \citep{ridnik2021asymmetric,huynh2020interactive}, we use both F1 scores and mAP as evaluation metrics in this task. Detailed methodology of the re-weight approach and the detailed formula of metric calculations can be found in Appendix \ref{appd:tech_details}.


\paragraph{Binary PU Learning}

We consider a 13-layer CNN with ReLU as the backbone and Adam as the optimizer. 
\citet{kiryo2017positive} designed an algorithm \textbf{nnPU} for PU learning with balanced binary classification data, while \citet{su2021positive} proposed \textbf{ImbalancednnPU} considering imbalanced setting. We take these two previous state-of-the-art models as our compared baselines. We rerun \textbf{nnPU} and \textbf{ImbalancennPU} with their provided codes and configurations and report the results. The negative reward sampling is $20\%$ for all settings. The threshold $\gamma$ to choose enhancement labels is $0.8$. We keep the values of other hyper-parameters the same as \citet{su2021positive}. Following previous work, we evaluate all the methods with F1 scores. Unless stated otherwise, the hyper-parameters specified in our framework remain the same in the following experiments.

\section{More experiments}
\label{appd:more_exp}

We indeed considered two different scenarios to evaluate the robustness of our method.

Different dataset versions: To simulate partial annotation, we randomly reserve a ratio of positive classes and treat other classes as unknown. To fully verify the effectiveness and stability of our model, we randomly constructed three versions of data sets (refer to \Cref{tab:ns_com} for detailed results on DocRE). It can be seen that our method achieves consistent improvement in all the dataset versions.
Different runs in the same dataset: For multi-label image classification, we repeat three runs in the same dataset. The standard deviations of F1/MAP scores of \Cref{tab:std} in the paper are as follows.

\begin{table*}[t!]
\setlength\tabcolsep{2pt}
\small
\centering
\resizebox{\textwidth}{!}{%
\begin{tabular}{c|lll |lll|lll|lll}
\toprule
 & \multicolumn{3}{c|}{\textbf{version0} }                                                     & \multicolumn{3}{c|}{\textbf{version1} }     & \multicolumn{3}{c|}{\textbf{version2} }              & \multicolumn{3}{c}{\textbf{average} }                         \\ 
     {\textbf{Sampling Ratio}}   & {P} & {R} & {F1} & {P} & {R} & {F1} & {P} & {R} & {F1} & {P} & {R} & {F1}\\ \midrule
 \textbf{0.1} & 71.21  &  82.27 & 76.34 & 77.19 & 77.3 & 77.25 & 81.53 & 71.14 & 75.98 & 76.64 & 76.9 & 76.52
\\
\textbf{0.2} &  75.22  & 79.23  & 77.17 & 79.72 & 74.82 & 77.19 & 83.26 & 68.4 & 75.14 & 79.4 & 74.15 & 76.5
\\ 
\textbf{0.3} &  76.79  & 77.88 & 77.33 & 80.57 & 73.6 & 76.93 & 84.88 & 66.84 & 74.79 & 80.75 & 72.77 & 76.35
\\
\textbf{0.4} &  78.26  & 77.18 & 77.72 & 81.1 & 73.58 & 77.16 & 83.94 & 67.4 & 74.77 & 81.1 & 72.72 & 76.55
\\
\textbf{0.5} &  78.35  &  77.1 & 77.72 & 83.26 & 72.1 & 77.28 & 85.23 & 65.7 & 74.2 & 82.28 & 71.63 & 76.4
\\
\textbf{0.6} &  79.54  &  76.04 & 77.75 & 83.22 & 71.41 & 76.86 & 85.41 & 65.74 & 74.29 & 82.72 & 71.06 & 76.3
\\
\textbf{0.7} &  80.74  &  75.35 & 77.95 & 83.03 & 71.12 & 76.61 & 86.41 & 64.4 & 73.8 & 83.39 & 70.29 & 76.12
\\
\textbf{0.8} &  80.43  &  75.12 & 77.68 & 83.92 & 70.6 & 76.69 & 84.65 & 65.7 & 73.98 & 83.0 & 70.47 & 76.12 
\\
\textbf{0.9} &  81.04  &  74.62 & 77.7 & 84.45 & 69.83 & 76.45 & 84.44 & 65.74 & 73.92 & 83.31 & 70.06 & 76.02
\\
\bottomrule
\end{tabular}
}
\caption{Action Sampling Ratio}
\label{tab:ns_com}

\end{table*}

\subsection{Document-level Relation Extraction}
\label{appd:doc_exp}

\begin{figure}[t!]
    \centering
    \includegraphics[width=\linewidth]{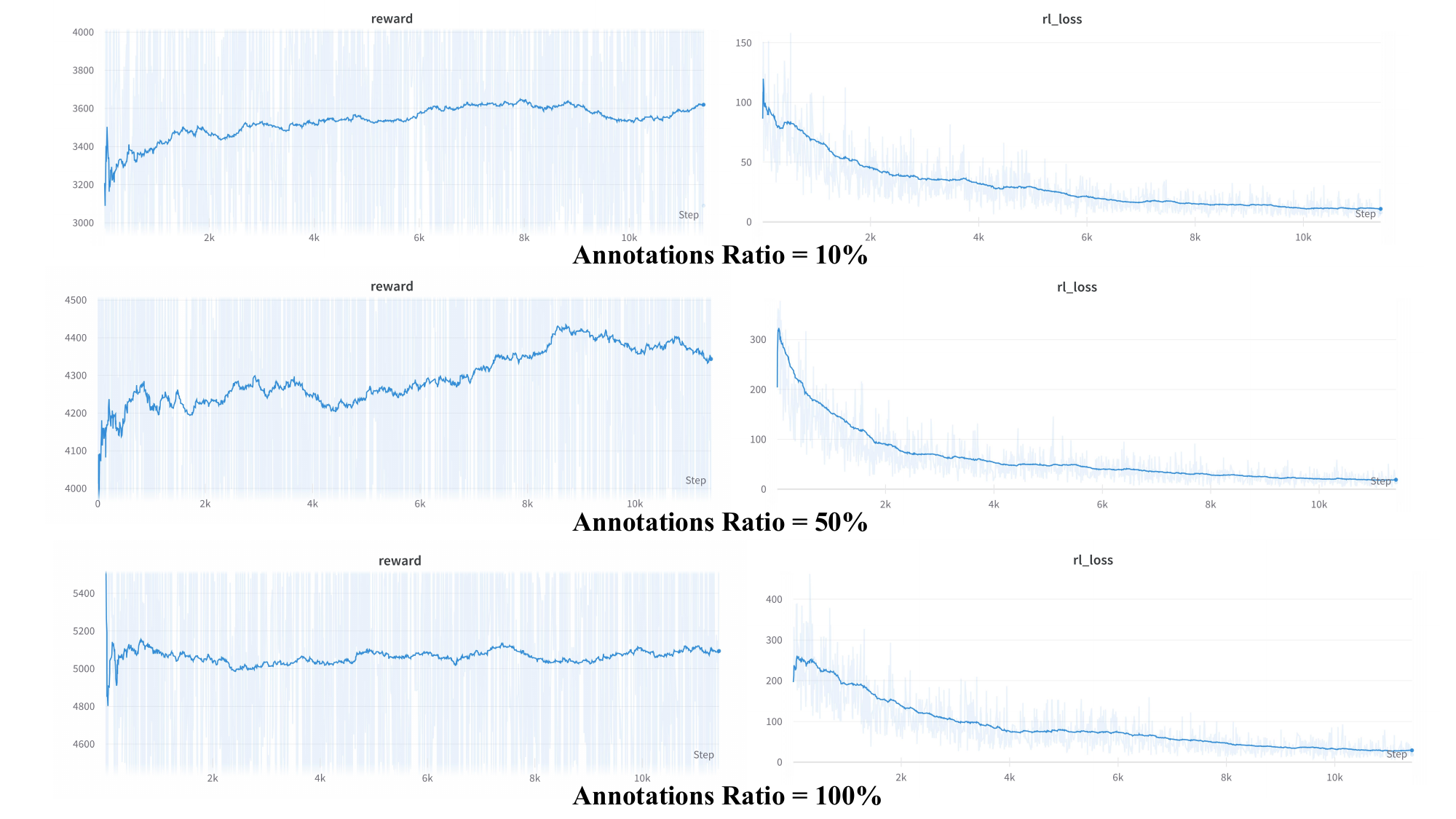}
    \caption{Train Curve.}
    \label{fig:train_curve}
\end{figure}

\textbf{Training curve}. In \cref{fig:train_curve}, we display the reward and loss curves of our model in three annotation rations, $10\%$, $50\%$, and $100\%$. Our experimental settings were conducted under partially annotated multi-label tasks, but we also compute metrics on ground Truth during the experiment. 

\textbf{All experiments on different ratios of annotated labels}
To fully verify the effectiveness and robustness of our model, we randomly constructed three versions of data sets and tested the DREEAM model, Pos Weight, Neg Weight, and our MLPAC model on all data sets respectively. The results are shown in \cref{tab:model_com_re_all}.

\textbf{Experiments on selecting action sampling ratios} (Take annotations ratio=50\% as an example) In order to select the action sampling ratio hyperparameter, we conducted comparative experiments from 0.1 to 0.9, and finally found that the model performed best when the hyperparameter was 0.4. The results are shown in \cref{tab:ns_com}.

\textbf{Critic network performance of Our MLPAC}
We iteratively train our critic network and policy network. After multiple rounds of iterations, the performance of the critic network has been greatly improved. The performance of the critic network of our MLPAC is shown in \cref{tab:value_per}.

\textbf{Case study}. In \cref{tab:re_example}, we show an example on the prediction of each method. Our MLPAC predicts more true positives.

\subsection{Multi-label Image Classification}
\label{appd:img_exp}

\textbf{The experimental results on extra evaluation metrics and annotation ratios}. In \cref{tab:toy_P_R} and \cref{tab:coco_extra}, we show Precision and Recall of CIFAR10 and the results of other annotation ratios on Ms-COCO. \cref{tab:std} shows the stability of our method MLPAC. The standard deviation was computed from three different runs on the MS-COCO dataset.

\textbf{The experimental results on synthetic classification.} According to the data construction, any category of data in CIFAR10 dataset can be chosen as the positive set. To further make our experiments convincing, we show the results of different data construction in \cref{tab:cifar10}.

\begin{table*}[h]
\setlength\tabcolsep{2pt}
\small
\centering
\resizebox{\textwidth}{!}{%
\begin{tabular}{c|l|lll|lll|lll|lll}
\toprule
 & & \multicolumn{3}{c|}{\textbf{version0} }                                                     & \multicolumn{3}{c|}{\textbf{version1} }     & \multicolumn{3}{c|}{\textbf{version2} }              & \multicolumn{3}{c}{\textbf{average} }                         \\ 
{\textbf{Method}}              &      {\textbf{Data Ratio}}   & {P} & {R} & {F1} & {P} & {R} & {F1} & {P} & {R} & {F1} & {P} & {R} & {F1}\\ \midrule
\multirow{9}{*}{DREEAM}  & \textbf{10\%} & 91.23  &  4.53 & 8.64 & 90.88 & 3.54 & 6.82 & 87.39 & 3.38 & 6.5 & 89.83 & 3.82 & 7.32
\\
& \textbf{20\%} &  90.74  & 10.0  & 18.02 & 92.37 & 9.65 & 17.47 & 90.86 & 9.8 & 17.69 & 91.32 & 9.82 & 17.73
\\ 
& \textbf{30\%} &  92.45  & 19.94 & 32.8 & 92.48 & 19.95 & 32.82 & 91.19 & 22.35 & 35.9 & 92.04 & 20.75 & 33.84
\\
& \textbf{40\%} &  93.12  & 28.08 & 43.15 & 91.92 & 34.82 & 50.51 & 92.25 & 36.34 & 52.14 & 92.43 & 33.08 & 48.6
\\
& \textbf{50\%} &  92.69  &  43.63 & 59.33 & 91.25 & 43.28 & 58.71 & 91.44 & 39.29 & 54.96 & 91.79 & 42.07 & 57.67
\\
& \textbf{60\%} &  92.16  &  48.56 & 63.6 & 90.49 & 56.52 & 69.58 & 89.52 & 55.68 & 68.66 & 90.72 & 53.59 & 67.28
\\
& \textbf{70\%} &  88.56  &  60.15 & 71.64 & 91.28 & 55.69 & 69.17 & 88.92 & 60.51 & 71.01 & 89.59 & 58.78 & 70.61
\\
& \textbf{80\%} &  87.91  &  65.18 & 74.86 & 89.83 & 62.75 & 73.89 & 86.95 & 66.49 & 75.36 & 88.23 & 64.81 & 74.7 
\\
& \textbf{90\%} &  87.49  &  66.86 & 75.79 & 87.61 & 67.66 & 76.35 & 86.4 & 68.75 & 76.57 & 87.17 & 67.76 & 76.24
\\
\midrule

\multirow{9}{*}{Pos Weight}  
& \textbf{10\%} & 84.43  & 34.1 & 48.57 & 84.61 & 43.22 & 57.21 & 85.8 & 42.21 & 56.58 & 84.95 & 39.84 & 54.12
\\
& \textbf{20\%} & 87.72  & 47.36 & 61.51 & 82.51 & 57.19 & 67.56 & 86.61 & 51.3 & 64.44 & 85.61 & 51.95 & 64.5
\\ 
& \textbf{30\%} & 83.57  & 61.65 & 70.95 & 87.05 & 57.04 & 68.92 & 85.75 & 59.23 & 70.07 & 85.46 & 59.31 & 69.98
\\
& \textbf{40\%} & 87.51 & 59.26 & 70.67 & 84.29 & 65.91 & 73.97 & 85.65 & 64.14 & 73.35 & 85.82 & 63.1 & 72.66
\\
& \textbf{50\%} & 83.66  & 68.09 & 75.08 & 85.78 & 66.33 & 74.81 & 85.66 & 65.92 & 74.5 & 85.03 & 66.78 & 74.8
\\
& \textbf{60\%} & 84.85 & 68.57 & 75.85 & 85.55 & 68.09 & 75.83 & 84.51 & 68.87 & 75.89 & 84.97 & 68.51 & 75.86
\\
& \textbf{70\%} & 82.77 & 73.07 & 77.62 & 83.0 & 73.13 & 77.76 & 84.37 & 71.4 & 77.34 & 83.38 & 72.53 & 77.57
\\
& \textbf{80\%} & 83.57  & 73.82 & 78.39 & 82.46 & 75.68 & 78.93 & 83.64 & 73.61 & 78.31 & 83.22 & 74.37 & 78.54
\\
& \textbf{90\%} & 83.9  & 74.54 & 78.94 & 82.48 & 76.44 & 79.35 & 82.87 & 75.8 & 79.18 & 83.08 & 75.59 & 79.16
\\
\midrule

\multirow{9}{*}{Neg Weight} 
& \textbf{10\%} & 88.1  & 29.67 & 44.39 & 86.06 & 30.1 & 44.6 & 86.06 & 32.37 & 47.05 & 86.74 & 30.71 & 45.35
\\ 
& \textbf{20\%} & 82.94 & 55.7 & 66.64 & 83.72 & 55.24 & 66.56 & 85.49 & 51.25 & 64.08 & 84.05 & 54.06 & 65.76
\\
& \textbf{30\%} & 85.9  & 58.87 & 69.86 & 86.47 & 55.99 & 67.97 & 82.7 & 63.04 & 71.55 & 85.02 & 59.3 & 69.79
\\
& \textbf{40\%} & 86.1  & 62.08 & 72.14 & 85.55 & 62.72 & 72.37 & 85.19 & 64.47 & 73.39 & 85.61 & 63.09 & 72.63
\\
& \textbf{50\%} & 84.25 & 67.83 & 75.15 & 84.27 & 68.5 & 75.57 & 83.64 & 68.37 & 75.24 & 84.05 & 68.23 & 75.32
\\
& \textbf{60\%} & 84.25 & 69.76 & 76.32 & 84.39 & 69.17 & 76.02 & 82.92 & 71.29 & 76.67 & 83.85 & 70.07 & 76.34
\\
& \textbf{70\%} & 81.89 & 73.52 & 77.48 & 82.88 & 73.03 & 77.64 & 83.74 & 71.72 & 77.27 & 82.84 & 72.76 & 77.46
\\
& \textbf{80\%} & 80.99  & 76.24 & 78.54 & 82.51 & 74.83 & 78.48 & 81.58 & 74.93 & 78.11 & 81.69 & 75.33 & 78.38
\\
& \textbf{90\%} & 80.85  & 77.08 & 78.92 & 80.7 & 76.93 & 78.77 & 80.92 & 77.22 & 79.03 & 80.82 & 77.08 & 78.91
\\
\midrule

\multirow{9}{*}{Our MLPAC}  
& \textbf{10\%} & 64.47 & 72.78 & 68.37 & 62.39 & 74.98 & 68.11 & 48.65 & 83.15 & 61.39 & 58.5 & 76.97 & 65.96
\\
& \textbf{20\%} & 82.25  & 66.75 & 73.69 & 86.2 & 58.91 & 69.99 & 81.94 & 67.33 & 73.92 & 83.46 & 64.33 & 72.53
\\
& \textbf{30\%} & 83.71  & 67.98 & 75.03 & 86.03 & 63.58 & 73.12 & 80.87 & 71.56 & 75.93 & 83.54 & 67.71 & 74.69
\\
& \textbf{40\%} & 84.56  & 68.92 & 75.94 & 83.21 & 70.08 & 76.08 & 83.78 & 69.2 & 75.8 & 83.85 & 69.4 & 75.94
\\
& \textbf{50\%} & 81.4  & 72.86 & 76.89 & 80.32 & 74.34 & 77.21 & 82.55 & 73.62 & 77.83 & 81.42 & 73.61 & 77.31
\\
& \textbf{60\%} & 82.3  & 73.97 & 77.92 & 80.4 & 75.35 & 77.79 & 80.61 & 74.7 & 77.54 & 81.1 & 74.67 & 77.75
\\
& \textbf{70\%} & 83.34  & 73.57 & 78.15 & 83.27 & 73.87 & 78.29 & 83.25 & 74.36 & 78.55 & 83.29 & 73.93 & 78.33
\\
& \textbf{80\%} & 81.92  & 75.65 & 78.66 & 81.68 & 76.02 & 78.75 & 62.77 & 80.57 & 70.56 & 75.46 & 77.41 & 75.99
\\
& \textbf{90\%} & 80.83  & 77.58 & 79.18 & 80.41 & 78.01 & 79.19 & 80.97 & 77.48 & 79.19 & 80.74 & 77.69 & 79.2
\\
\bottomrule
\end{tabular}
}
\caption{Results of DREEAM, Pos Weight, Neg Weight, MLPAC on different ratios of annotated labels}
\label{tab:model_com_re_all}

\end{table*}

\begin{table*}[t!]
\setlength\tabcolsep{2pt}
\small
\centering
\resizebox{\textwidth}{!}{%
\begin{tabular}{c|lll |lll|lll|lll}
\toprule
  & \multicolumn{3}{c|}{\textbf{version0} }                                                     & \multicolumn{3}{c|}{\textbf{version1} }     & \multicolumn{3}{c|}{\textbf{version2} }              & \multicolumn{3}{c}{\textbf{average} }                         \\ 
      {\textbf{Data Ratio}}   & {P} & {R} & {F1} & {P} & {R} & {F1} & {P} & {R} & {F1} & {P} & {R} & {F1}\\ \midrule
10\% & 60.69 & 74.91 & 67.06 & 57.47 & 77.17 & 65.88 & 45.89 & 84.41 & 59.46 & 54.68 & 78.83       & 64.13 \\
20\% & 81.16 & 67.2  & 73.52 & 86.34 & 58.12 & 69.47 & 83.3  & 63.88 & 72.3  & 83.6        & 63.07 & 71.76 \\
30\% & 83.1  & 66.99 & 74.18 & 83.65 & 64.82 & 73.04 & 80.95 & 69.74 & 74.93 & 82.57 & 67.18 & 74.05       \\
40\% & 85.19 & 66.35 & 74.6  & 83.25 & 69.41 & 75.7  & 83.12 & 69.13 & 75.48 & 83.85 & 68.3 & 75.26       \\
50\% & 80.44 & 72.78 & 76.42 & 78.51 & 75.1  & 76.76 & 83.24 & 72.89 & 77.28 & 80.73       & 73.59       & 76.82       \\
60\% & 83.53 & 71.8  & 77.22 & 79.9  & 74.35 & 77.02 & 80.59 & 73.6  & 76.93 & 81.34       & 73.25       & 77.06 \\
70\% & 86.14 & 69.18 & 76.74 & 87.65 & 66.63 & 75.71 & 85.57 & 69.9  & 76.94 & 86.45 & 68.57       & 76.46 \\
80\% & 85.77 & 70.19 & 77.2  & 83.64 & 72.85 & 77.87 & 79.93 & 71.16 & 75.29 & 83.11 & 71.4        & 76.79 \\
90\% & 83.86 & 74.78 & 79.06 & 82.85 & 74.42 & 78.41 & 83.77 & 74.42 & 78.82 & 83.49 & 74.54       & 78.76
\\
\bottomrule
\end{tabular}
}
\caption{Critic network performance of Our MLPAC. We construct the training set three times with different random seeds, corresponding to the three versions.}
\label{tab:value_per}

\end{table*}

\begin{table*}[t!]
\setlength\tabcolsep{2pt}
\scriptsize
\centering
\resizebox{0.92\textwidth}{!}{%
\begin{tabular}{c|lll |lll|lll|lll}
\toprule
  & \multicolumn{3}{c|}{\textbf{nnPU} }                                                     & \multicolumn{3}{c|}{\textbf{ImbnnPU} }     & \multicolumn{3}{c|}{\textbf{Negative Mode} }              & \multicolumn{3}{c}{\textbf{Our MLPAC} }                         \\ 
      {\textbf{Data Ratio}}   & {P} & {R} & {F1} & {P} & {R} & {F1} & {P} & {R} & {F1} & {P} & {R} & {F1}\\ \midrule
10\%& 52.0& 39.3& 44.8& 41.3& 59.2& 48.6& 40.4& 3.8& 7.0& 47.7& 53.0& 50.2\\
20\%& 54.6& 42.6& 47.9& 43.9& 66.8& 53.0& 76.8& 9.6& 17.1& 61.4& 68.1& 64.6\\
30\%& 58.4& 42.3& 49.1& 43.9& 66.8& 53.0& 71.1& 18.0& 28.7& 59.6& 75.6& 66.6\\
40\%& 57.1& 45.1& 50.4& 54.8& 73.7& 62.8& 76.9& 24.9& 37.6& 62.1& 74.6& 67.8\\
50\%& 56.9& 49.2& 52.8& 61.5& 69.4& 65.2& 75.0& 45.8& 56.9& 63.8& 76.9& 69.8\\
60\%& 59.4& 49.7& 54.1& 61.5& 68.1& 64.6& 69.1& 46.5& 55.6& 65.0& 77.7& 70.8\\
70\%& 61.7& 51.5& 56.1& 62.6& 67.4& 64.9& 82.2& 52.7& 64.2& 72.6& 77.7& 75.1\\
80\%& 63.0& 52.7& 57.4& 63.2& 72.8& 67.7& 79.5& 64.4& 71.2& 78.9& 73.0& 75.8\\
90\%& 70.4& 47.5& 56.7& 62.7& 77.2& 69.2& 82.5& 69.1& 75.2& 75.2& 78.7& 76.9\\
\bottomrule
\end{tabular}
}
\caption{The results of CIFAR10 dataset. We consider the original class `airplane' as the positive targets.}
\label{tab:toy_P_R}
\end{table*}

\begin{table*}[t!]
\setlength\tabcolsep{2pt}
\small
\centering
\resizebox{0.92\textwidth}{!}{%
\begin{tabular}{c|lll |lll|lll|lll}
\toprule
  & \multicolumn{3}{c|}{\textbf{Pos Weight} }                                                     & \multicolumn{3}{c|}{\textbf{Neg Weight} }     & \multicolumn{3}{c|}{\textbf{Negative Mode} }              & \multicolumn{3}{c}{\textbf{Our MLPAC} }                         \\ 
      {\textbf{Data Ratio}}   & {P} & {R} & {F1} & {P} & {R} & {F1} & {P} & {R} & {F1} & {P} & {R} & {F1}\\ \midrule
20\%& 72.8& 69.5& 71.1& 89.7& 39.5& 54.9& 87.9& 9.9& 17.9& 79.4& 67.2& 72.8\\
40\%& 74.3& 75.0& 74.7& 82.5& 65.6& 73.1& 90.3& 28.0& 42.8& 83.0& 74.4& 78.5\\
60\%& 74.6& 79.0& 76.7& 80.1& 74.0& 76.9& 96.0& 47.0& 63.1& 79.4& 76.5& 77.9\\
80\%& 72.5& 83.1& 77.4& 83.1& 75.7& 79.2& 92.8& 65.3& 76.6& 82.4& 77.5& 79.9\\
\bottomrule
\end{tabular}
}
\caption{The results of other annotation ratios on MS-COCO dataset.}
\label{tab:coco_extra}
\end{table*}

\begin{table*}[t!]
\centering
\large
\resizebox{\textwidth}{!}{
\begin{tabular}{cc|cc|cc|cc|cc}
\toprule
\multicolumn{10}{c}{\textbf{Standard Deviation} }\\
\midrule
                 \multicolumn{2}{c|}{\textbf{10\%} }                                                     & \multicolumn{2}{c|}{\textbf{30\%} }     & \multicolumn{2}{c|}{\textbf{50\%} }
                & \multicolumn{2}{c|}{\textbf{70\%} }                                       & \multicolumn{2}{c}{\textbf{90\%}}                                       \\ 
        {F1} & {mAP} & {F1} & {mAP} & {F1} & {mAP} & {F1} & {mAP}& {F1} & {mAP}\\ \midrule
 68.3(0.12) & 66.6(0.33) & 77.0(0.30) & 77.5(0.25) & 79.1(0.15) & 80.4(0.13) & 79.0(0.10) & 81.4(0.15) & 80.5(0.05) & 83.4(0.05) \\

\bottomrule
\end{tabular}
}
\caption{The standard deviations ($\cdot$) of F1 and mAP were computed from three different runs on the MS-COCO dataset.}
\label{tab:std}
\end{table*}

\begin{table*}[t!]
\small
\centering

\label{tab:cifar10}
\begin{tabular}{l|lllllllll}
\toprule
{\textbf{Method}}   & \textbf{10\%} & \textbf{20\%} & \textbf{30\%} & \textbf{40\%} & \textbf{50\%} & \textbf{60\%} & \textbf{70\%} & \textbf{80\%} & \textbf{90\%}\\ 
\midrule
\textbf{nnPU} & 44.8 & 47.9 & 49.1 & 50.4 & 52.8 & 54.1 & 56.1 & 57.4 & 56.7 \\
\textbf{ImbalancednnPU} & 48.6  & 53.0  & 59.1  & 62.8  & 65.2  & 64.6 & 64.9  & 67.7  & 69.2 \\
\midrule
\textbf{Negative Mode} & 7.0  & 17.1  & 28.7 & 37.6 & 56.9 & 55.6 & 64.2 & 71.2 & 75.2 \\ 
\textbf{MLPAC~(Ours)} & 50.2  & 64.6 & 66.6 & 67.8 & 69.8 & 70.8 & 75.1 & 75.8 & 76.9 \\
\bottomrule
\end{tabular}
\caption{F1 scores with varying ratios of positive annotations. We take images of the ``Airplane" category as positives in this table.}
\end{table*}

\begin{table*}
\small
\begin{tabular}{l|p{10.8cm}}
\toprule
\textbf{Item} & \textbf{Content or Triples} \\ \midrule
\textbf{Title} & {Guido Bonatti} \\ \midrule
\textbf{Document} & {Guido Bonatti (died between 1296 and 1300) was an Italian mathematician, astronomer and astrologer, who was the most celebrated astrologer of the 13th century. Bonatti was advisor of Frederick II, Holy Roman Emperor, Ezzelino da Romano III, Guido Novello da Polenta and Guido I da Montefeltro. He also served the communal governments of Florence, Siena and Forlì. His employers were all Ghibellines (supporters of the Holy Roman Emperor), who were in conflict with the Guelphs (supporters of the Pope), and all were excommunicated at some time or another. Bonatti 's astrological reputation was also criticised in Dante's Divine Comedy, where he is depicted as residing in hell as punishment for his astrology. His most famous work was his Liber Astronomiae or 'Book of Astronomy', written around 1277. This remained a classic astrology textbook for two centuries.} \\ \midrule
\textbf{DREEAM} & {${\langle}$Dante, notable work,  Divine Comedy${\rangle}$} \\  \midrule

\multirow{3}{*}{\textbf{Pos Weight}} & {${\langle}$Dante, notable work,  Divine Comedy${\rangle}$} \\
& {${\langle}$Divine Comedy, creator,  Dante${\rangle}$} \\
& {${\langle}$Divine Comedy, author,  Dante${\rangle}$} \\  \midrule

\multirow{4}{*}{\textbf{Neg Weight}} 
& {${\langle}$Guido Bonatti, notable work,  Liber Astronomiae${\rangle}$} \\ 
& {${\langle}$Guido Bonatti, notable work,  Book of Astronomy${\rangle}$} \\ 
& {${\langle}$Dante, notable work,  Divine Comedy${\rangle}$} \\ 
& {${\langle}$Divine Comedy, author,  Dante${\rangle}$} \\ \midrule

\multirow{6}{*}{\textbf{Our MLPAC}}  
& {${\langle}$Guido Bonatti, notable work,  Liber Astronomiae${\rangle}$} \\
& {${\langle}$Guido Bonatti, notable work,  Book of Astronomy${\rangle}$} \\
& {${\langle}$Dante, notable work,  Divine Comedy${\rangle}$} \\
& {${\langle}$Divine Comedy, creator,  Dante${\rangle}$} \\
& {${\langle}$Divine Comedy, author,  Dante${\rangle}$} \\
& {${\langle}$Liber Astronomiae, author,  Guido Bonatti${\rangle}$} \\ \midrule

\multirow{12}{*}{\textbf{Ground Truth}}   
& {${\langle}$Guido Bonatti, date of death,  1296${\rangle}$} \\
& {${\langle}$Guido Bonatti, date of death,  1300${\rangle}$} \\
& {${\langle}$Divine Comedy, characters,  Guido Bonatti${\rangle}$} \\
& {${\langle}$Divine Comedy, creator,  Dante${\rangle}$} \\
& {${\langle}$Divine Comedy, author,  Dante${\rangle}$} \\
& {${\langle}$Book of Astronomy, author,  Guido Bonatti${\rangle}$} \\
& {${\langle}$Liber Astronomiae, author,  Guido Bonatti${\rangle}$} \\
& {${\langle}$Guido Bonatti, country of citizenship,  Italian${\rangle}$} \\
& {${\langle}$Guido Bonatti, notable work,  Liber Astronomiae${\rangle}$} \\
& {${\langle}$Dante, notable work,  Divine Comedy${\rangle}$} \\
& {${\langle}$Guido Bonatti, present in work,  Divine Comedy${\rangle}$} \\
& {${\langle}$Guido Bonatti, notable work,  Book of Astronomy${\rangle}$}
\\ \bottomrule
\end{tabular}
\caption{An Example from Re-DocRED}
\label{tab:re_example}
\end{table*}



\end{document}